\documentclass[sigconf]{acmart}

\copyrightyear{2025}
\acmYear{2025}
\setcopyright{acmlicensed}\acmConference[MM '25]{Proceedings of the 33rd ACM International Conference on Multimedia}{October 27--31, 2025}{Dublin, Ireland}
\acmBooktitle{Proceedings of the 33rd ACM International Conference on Multimedia (MM '25), October 27--31, 2025, Dublin, Ireland}
\acmDOI{10.1145/3746027.3755317}
\acmISBN{979-8-4007-2035-2/2025/10}

\usepackage{booktabs}
\usepackage{multirow}
\usepackage{balance}
\begin{document}

\title{HGC-Avatar: Hierarchical Gaussian Compression for Streamable Dynamic 3D Avatars}

\settopmatter{authorsperrow=4}

\author{Haocheng Tang}
\affiliation{
 \institution{State Key Laboratory of Multimedia Information Processing, School of Computer Science, Peking University.}
 \city{Beijing}
 \country{China}
}
\email{hctang@stu.pku.edu.cn}

\author{Ruoke Yan}
\affiliation{
 \institution{State Key Laboratory of Multimedia Information Processing, School of Computer Science, Peking University.}
 \city{Beijing}
 \country{China}
}
\email{ruoke.yan@stu.pku.edu.cn}

\author{Xinhui Yin}
\affiliation{
 \institution{State Key Laboratory of Multimedia Information Processing, School of Computer Science, Peking University.}
\city{Beijing}
 \country{China}
}
\email{yinxh23@mails.jlu.edu.cn}

\author{Qi Zhang}
\affiliation{
 \institution{State Key Laboratory of Multimedia Information Processing, School of Computer Science, Peking University.}
  \city{Beijing}
 \country{China}
}
\email{ywwynm@pku.edu.cn}

\author{Xinfeng Zhang}
\affiliation{
  \institution{School of Computer Science and Technology, University of Chinese Academy of Sciences.}
  \city{Beijing}
  \country{China}
 }
 \email{xfzhang@ucas.ac.cn}

\author{Siwei Ma}
\affiliation{
 \institution{State Key Laboratory of Multimedia Information Processing, School of Computer Science, Peking University.}
  \city{Beijing}
 \country{China}
}
\email{swma@pku.edu.cn}

\author{Wen Gao}
\affiliation{
 \institution{State Key Laboratory of Multimedia Information Processing, School of Computer Science, Peking University.}
  \city{Beijing}
 \country{China}
}
\email{wgao@pku.edu.cn}

\author{Chuanmin Jia}
\authornote{H. Tang and R. Yan contributed equally. Corresponding author: S. Ma and C. Jia }
\affiliation{
  \institution{WICT, State Key Laboratory of Multimedia Information Processing, Peking University.}
\city{Beijing}
 \country{China}
}
\email{cmjia@pku.edu.cn}

\renewcommand{\shortauthors}{Haocheng Tang et al.}

\begin{teaserfigure}
  \includegraphics[width=\textwidth]{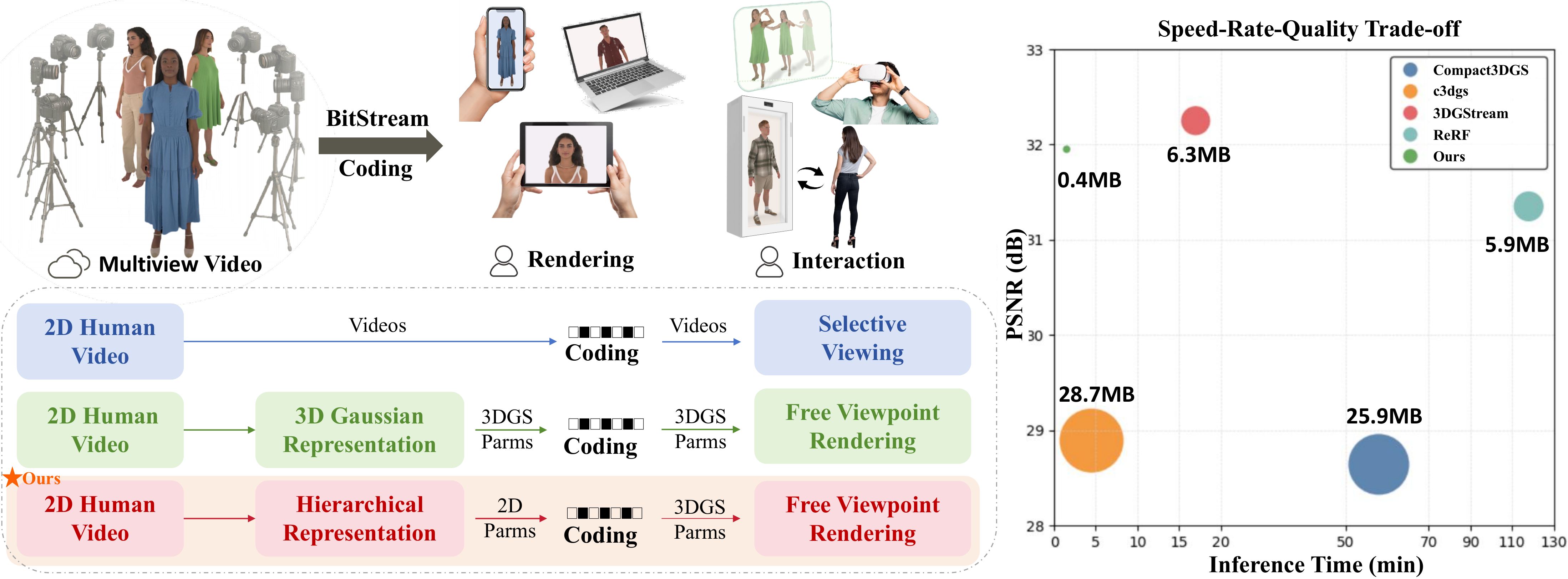}
  \caption{We propose an efficient encoding scheme for dynamic 3D avatars, enabling cloud-based human compression and high-fidelity rendering on the application side, ideal for immersive telepresence and interactive applications. Compared to traditional and Gaussian compression methods, our approach excels in parameter encoding for human representation, delivering superior performance in speed, bitrate, and reconstruction quality.}
  \label{fig:teaser}
\end{teaserfigure}

\begin{abstract}
Recent advances in 3D Gaussian Splatting (3DGS) have enabled fast, photorealistic rendering of dynamic 3D scenes, showing strong potential in immersive communication. However, in digital human encoding and transmission, the compression methods based on general 3DGS representations are limited by the lack of human priors, resulting in suboptimal bitrate efficiency and reconstruction quality at the decoder side, which hinders their application in streamable 3D avatar systems. We propose HGC-Avatar, a novel Hierarchical Gaussian Compression framework designed for efficient transmission and high-quality rendering of dynamic avatars. Our method disentangles the Gaussian representation into a structural layer, which maps poses to Gaussians via a StyleUNet-based generator, and a motion layer, which leverages the SMPL-X model to represent temporal pose variations compactly and semantically. This hierarchical design supports layer-wise compression, progressive decoding, and controllable rendering from diverse pose inputs such as video sequences or text. Since people are most concerned with facial realism, we incorporate a facial attention mechanism during StyleUNet training to preserve identity and expression details under low-bitrate constraints. Experimental results demonstrate that HGC-Avatar provides a streamable solution for rapid 3D avatar rendering, while significantly outperforming prior methods in both visual quality and compression efficiency.
\end{abstract}

\ccsdesc[300]{Human-centered computing}
\ccsdesc[300]{Computing methodologies~Reconstruction}
\ccsdesc[100]{Computing methodologies~Animation}

\keywords{Gaussian-based Human Compression; Hierarchical Human Representation; Streamable 3D Avatar}

\maketitle

\section{INTRODUCTION}

Immersive media has emerged as a key frontier in multimedia technology development. In most immersive applications, faithful reconstruction, efficient transmission, and realistic rendering of high-fidelity dynamic digital humans constitute fundamental requirements to improve immersion and optimal user engagement (Figure~\ref{fig:teaser}). To ensure a high-quality user experience in immersive communication and conferencing, it is essential to design representation methods for dynamic digital humans that support accurate reconstruction, efficient compression, and fast rendering.

Existing dynamic digital human representations fall into two categories (Figure~\ref{fig:teaser}). The first is traditional multi-view video, which leverages the well-established pipeline of 2D video acquisition, encoding, and display. However, it is constrained by the numbers and angles of available viewpoints, offering limited freedom in viewing and interaction. The second, more recent and popular representation is 3D Gaussian Splatting (3DGS)~\cite{kerbl20233Dgaussians}, which achieves high-quality reconstruction, efficient rendering, and supports free-viewpoint viewing and interaction. To support this representation, researchers have proposed a series of efficient compression methods~\cite{lee2024c3dgs, Niedermayr2024Compressed, sun20243dgstream} that enable compact 3DGS representations, significantly reducing transmission bandwidth and storage costs.

However, most existing 3DGS compression methods are designed for general-purpose scenarios and are not specifically optimized for dynamic digital humans, a unique form of 3D content. As a result, their compression efficiency is limited. Specifically, dynamic digital humans possess strong and exploitable prior characteristics. First, the human body exhibits a relatively stable and structurally similar form among individuals. Second, human motion can be viewed as deformations occurring while the underlying body structure remains unchanged. By leveraging these priors, compression can be further optimized to improve efficiency. In contrast, general 3DGS compression methods fail to take advantage of such human-centric information, resulting in suboptimal compression rates and reconstruction quality. Therefore, there is an urgent need for a compact and streamable representation of dynamic 3D avatars, enabling efficient rendering across platforms.

To enable efficient 3DGS representation and compression for digital humans, we propose \textbf{HGC-Avatar}, a hierarchical compression framework for streamable avatar rendering.
Based on the observation that such avatars exhibit temporal coherence, with per-frame variations primarily driven by pose, we design a two-layer hierarchical representation:
\textbf{(1) Motion layer}: Encodes temporal dynamics via SMPL-X pose parameters and generates corresponding pose maps, and \textbf{(2) Structural layer}: Stores a StyleUNet~\cite{kumar2024styleunet}-based generator that maps these pose maps to frame-specific Gaussian parameters, enabling geometry and appearance reconstruction without storing per-frame Gaussians. In addition, we introduce a facial attention module to enhance the capture of expression details, improving the perceptual quality of the face after decoding.

With this representation, compression is only required for 2D parameter information, including the SMPL-X parameters, the StyleUNet network, and pose maps. The hierarchical design allows independent encoding and decoding of structural and motion layers, supporting multi-modal pose inputs and controllable rendering on client devices. We extract poses based on user movement or generate poses from text prompts and present the specified pose of the reconstructed avatar at the decoding stage, enabling controllable interaction. As shown in Figure~\ref{fig:teaser}, our method significantly reduces data redundancy while maintaining high visual fidelity, and enables fast rendering on the decoder side. On widely used datasets like THuman4.0~\cite{zheng2022structured}, AvatarRex~\cite{zheng2023avatarrex} and ActorsHQ~\cite{icsik2023humanrf}, HGC-Avatar achieves an average PSNR of nearly 30dB with bitrate below 0.5MB per frame, surpassing SOTA Gaussian-based human reconstruction and Gaussian compression methods. These results demonstrate that HGC-Avatar is a practical and scalable solution for high-fidelity dynamic 3D human rendering on resource-limited platforms.
The main contributions of this paper can be summarized as follows:

\begin{itemize}
\item We introduce the first hierarchical Gaussian compression framework for dynamic 3D avatars, which disentangles structural and motion components to improve compression efficiency and supports controllable rendering from multi-modal pose inputs.

\item We use the SMPL-X model to encode human motion with parameterized poses, enabling compact, semantically meaningful motion modeling, which forms the basis for generating pose maps that guide the structural layer.

\item We integrate a facial attention module into the StyleUNet training, adaptively emphasizing facial regions during loss computation to enhance expression detail, especially in low-bitrate scenarios where perceptual quality is critical.

\item Extensive experiments show that our method achieves superior visual quality at low bitrates and enables efficient deployment on immersive applications like holographic cabins.
\end{itemize}

\section{RELATED WORK}
\subsection{Representation of 3D Humans}
Recent advances in 3D human representation evolved from explicit methods (meshes/point clouds) requiring dense inputs~\cite{Newcombe2011Kinect,Newcombe2015Dynamic} to neural approaches. Notably, Neural Radiance Fields (NeRF)~\cite{Mildenhall2021NeRF} achieved photorealistic synthesis via MLPs, later extended to dynamic scenes by D-NeRF~\cite{Pumarola2021DNerf}. For digital humans, Neural Body~\cite{peng2021neural} integrated SMPL priors into volumetric structures, while Animatable NeRF~\cite{peng2021animatable} pioneered deformation-field-based reconstruction. HumanNeRF~\cite{weng2022humannerf} further unified rigid/non-rigid deformation handling.
With the rise of 3D Gaussian Splatting (3DGS)~\cite{kerbl20233Dgaussians}, 3DGS-Avatar~\cite{qian20243dgs} inherits the geometric human modeling approach using both rigid and non-rigid deformations, then employs an MLP for color rendering. In addition, GaussianAvatar~\cite{hu2024gaussianavatar} introduces dynamic attributes to enable pose-dependent appearance modeling, while SplattingAvatar~\cite{shao2024splattingavatar} achieves the integration of trainable Gaussian embeddings with mesh representations. 4K4D~\cite{xu20244k4d} enables high-quality and fast novel view synthesis of dynamic humans through an efficient 4D point cloud representation.

\begin{figure*}[htp]
    \centering
    \includegraphics[width=17.5cm]{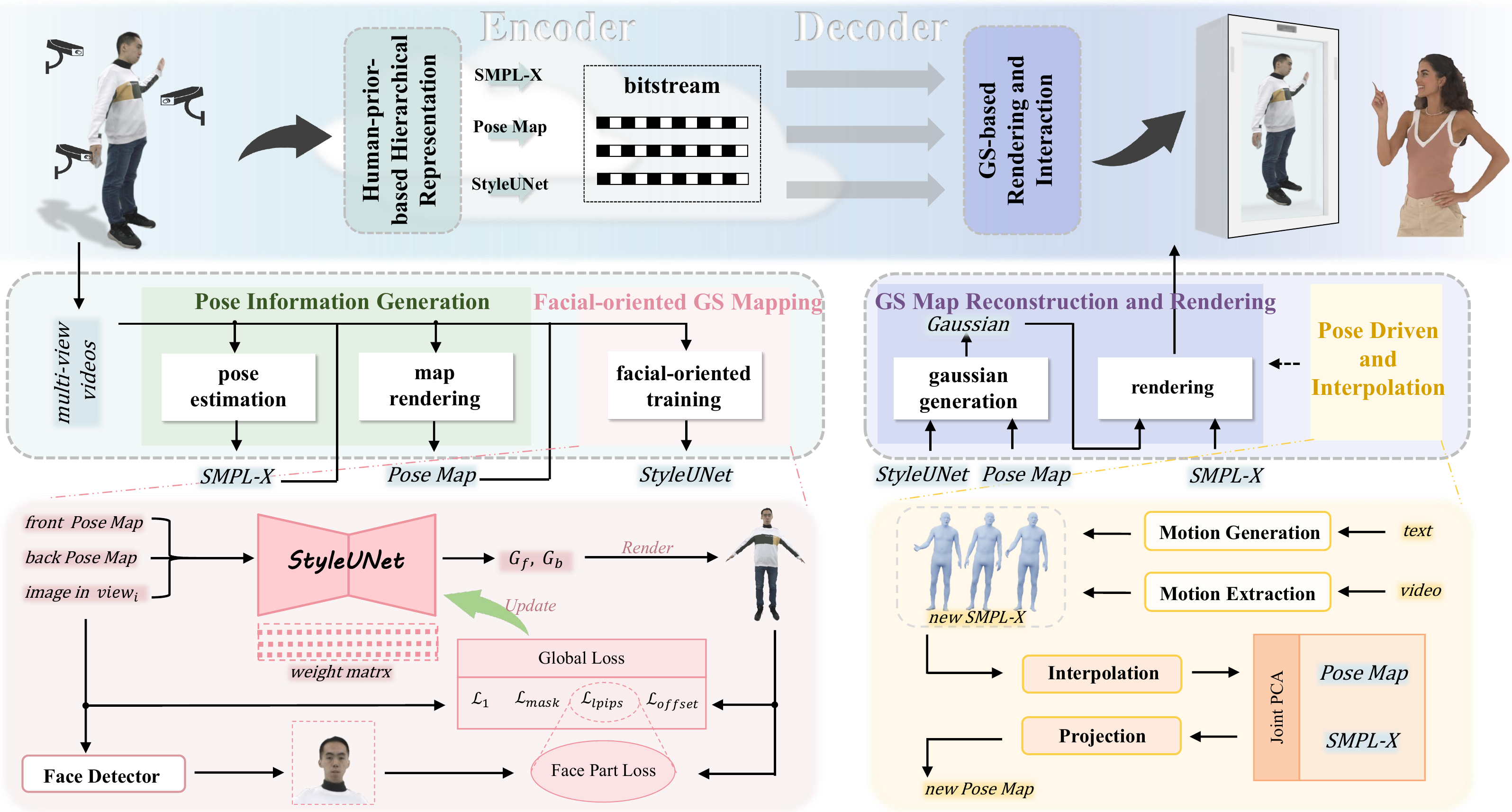}
    \caption{Overview of the proposed framework. Our method consists of three main components: human-prior-based representation, layered compression, and Gaussian-based rendering and interaction. It takes multi-view videos as input, extracts SMPL-X poses, renders pose maps, and uses a facial-oriented StyleUNet to generate Gaussian parameters. At the decoder side, Gaussian reconstruction enables low-latency, controllable avatar rendering driven by video or text inputs.}
    \label{fig:mainfig}
\end{figure*}

\subsection{3DGS Compression}

3DGS poses significant memory and storage challenges due to its many Gaussian parameters, leading to the development of various compression techniques. These can be divided into unstructured and structured methods. Unstructured techniques directly compress individual parameters, including pruning based on size, gradient, or opacity~\cite{lee2024c3dgs,fan2024lightgaussian}, quantization through vector or scalar methods~\cite{lee2024c3dgs,fan2024lightgaussian,Navaneet2024CompGS,Niedermayr2024Compressed}, and entropy coding to reduce redundancy~\cite{girish2024eagles}. Structured methods, on the other hand, utilize spatial and contextual relationships, employing anchor-based representations~\cite{lu2024scaffold,chen2024hac}, prediction models~\cite{wang2024contextgs,liu2024compgs}, graph-based structures~\cite{yang2024spectrally,zhang2024gaussianforest}, and tensor decomposition~\cite{sun2024f}. While structured methods offer better compression ratios and fidelity, they typically require more computational resources during rendering.

\subsection{Compression for 3D Humans}

Current 3D human compression methods fail to fully leverage human-specific properties, as human bodies exhibit stable geometry and temporally consistent attributes, with pose and motion as the primary variables. While pose-driven mesh compression achieves high geometric ratios~\cite{yan2023model,yan2024pose}, visual attributes like color/texture are often ignored. We propose incorporating human structure/motion knowledge into 3D Gaussian representations for joint geometry-attribute compression, offering efficient, render-friendly results balancing compression ratio, fidelity, and speed. Recent works like HiFi4G~\cite{jiang2024hifi4g}, DualGS~\cite{jiang2024robust}, and V3~\cite{wang2024v} focus on optimizing human representation using Gaussian-based methods for better compression. In contrast, VideoRF~\cite{wang2024videorf} and V3~\cite{wang2024v} are more application-focused, aiming to improve the practical use of these techniques in media. However, all these methods still rely on traditional Gaussian ellipsoids, which limits their compression performance.

\section{METHOD}
\subsection{Overview}
We propose a rendering-friendly compression framework for 3DGS-based digital humans, utilizing a hierarchical encoding-decoding strategy for efficient transmission and low-latency interaction. The framework, shown in Figure~\ref{fig:mainfig}, consists of three stages: human-prior-guided hierarchical representation, layered compression, and Gaussian-based rendering and reconstruction.

At the capture end, multi-view images and camera parameters are collected. SMPL-X parameters are extracted for pose information and used to generate pose maps, which are input to a StyleUNet trained to map them to frame-specific Gaussian parameters. We design compression schemes for pose parameters, pose maps, and StyleUNet weights, ensuring efficient encoding and transmission. On the decoder side, edge devices recover the StyleUNet~\cite{kumar2024styleunet} and pose data, generating Gaussian parameters for high-fidelity rendering. The framework also supports multi-modal pose control (e.g., video or text) for interactive applications. This hierarchical design enhances data compactness, transmission efficiency, and high-quality rendering in immersive scenarios.

\subsection{Human-prior-based Hierarchical Representation}
Inspired by the template-guided parameterization method Animatable Gaussians~\cite{li2024animatable}, we construct a hierarchical representation of dynamic avatars guided by human priors. SMPL-X pose parameters from multi-view videos are rendered into pose maps as structural cues for Gaussian generation. A StyleUNet maps these to frame-wise Gaussian parameters, with a facial attention module enhancing facial fidelity for improved identity/expression reconstruction. Overall, this stage consists of two components: pose information generation and facial-oriented Gaussian parameter mapping, as illustrated in the middle and bottom-left sections of Figure~\ref{fig:mainfig}.

\textbf{Pose Information Generation.}  
Accurate extraction of frame-wise pose information is fundamental to construct high-quality dynamic human representations. This process involves two main steps: (1) estimating SMPL-X parameters and (2) generating pose maps based on these estimates.

In the first step, SMPL-X~\cite{Pavlakos2019SMPL-X} is adopted as the parametric model for representing body pose $\boldsymbol{\theta}$, shape $\boldsymbol{\beta}$, and facial expression $\boldsymbol{\psi}$. Given multi-view images and camera parameters, these parameters are estimated by fitting in 2D domain~\cite{zhang2023pymaf}. SMPL-X extends SMPL by incorporating articulated hands and facial blendshapes, enabling full-body motion representation, which is defined as:
\begin{equation}
\mathcal{M} = \text{SMPL-X}(\boldsymbol{\theta}, \boldsymbol{\beta}, \boldsymbol{\psi}) \in \mathbb{R}^{N \times 3},
\end{equation}
where $\mathcal{M}$ denotes the mesh vertices. The fitting process minimizes the reprojection error between model joints and image observations, regularized by human priors.

In the second step, pose maps are generated using the estimated SMPL-X parameters, with keyframes near the canonical A-pose as reference templates. Signed distance fields (SDFs) and color fields are learned in the canonical space using implicit volumetric representations~\cite{Yariv2021Volume}. A 3D skinning weight volume is created by diffusing bone weights along the SMPL-X mesh normals to ensure consistent deformation between canonical and posed spaces.

Each canonical template is deformed to its pose using Linear Blend Skinning (LBS), preserving local motion by discarding global transformations. The posed mesh vertices are encoded as pseudo-color representations and orthographically projected to generate front and back view pose maps, which serve as compact, view-aligned inputs for the next module.

\textbf{Facial-oriented GS Parameter Mapping.}
To learn a reliable mapping from pose maps to per-frame Gaussian representations, we employ a convolutional neural network built upon the StyleGAN architecture, referred to as StyleUNet~\cite{wang2023styleavatar}. Given a single-frame 2D pose map, the network predicts the corresponding Gaussian parameters, including the spatial center, scale, and color of each Gaussian component. These parameters are subsequently used in a neural rendering pipeline to reconstruct high-fidelity 3D human appearances across diverse pose conditions. StyleUNet integrates multi-scale feature encoding, in U-Net, with progressive style modulation, in StyleGAN, enabling effective modeling of both global structure and local detail variations. The training of StyleUNet is guided by a composite objective that jointly optimizes image-level fidelity, geometric alignment, perceptual similarity, and spatial consistency. The overall loss function is defined as:
\begin{equation}
\mathcal{L}_{\text{total}} = 
w_{\text{L1}} \cdot \mathcal{L}_{\text{L1}} +
w_{\text{mask}} \cdot \mathcal{L}_{\text{mask}} +
w_{\text{lpips}} \cdot \mathcal{L}_{\text{lpips}} +
w_{\text{offset}} \cdot \mathcal{L}_{\text{offset}},
\end{equation}
where $\mathcal{L}_{\text{L1}}$ penalizes pixel-wise differences between the rendered output and ground-truth images, enhancing low-level reconstruction quality. $\mathcal{L}_{\text{mask}}$ encourages alignment between the predicted and ground-truth silhouettes. $\mathcal{L}_{\text{lpips}}$ enforces perceptual similarity in the deep feature space, with an emphasis on facial regions. $\mathcal{L}_{\text{offset}}$ regularizes the predicted Gaussian positions to suppress spatial drift and structural artifacts.

Considering the perceptual significance of facial regions in downstream tasks such as communication and expression synthesis, we introduce a Facial Attention Module to enhance the modeling of facial geometry and appearance. This module leverages spatial priors to explicitly guide the network's focus toward facial areas during training. Specifically, a binary face mask is employed to localize the target region, and a facial-aware perceptual loss is introduced to progressively increase the emphasis on facial reconstruction as training advances.

The perceptual loss is defined as:
\begin{equation}
\mathcal{L}_{\text{lpips}} = 
\sum_{k=1}^{L} \mathcal{A} \left(
W_k \cdot \left\| \mathbf{F}_k^{(0)} - \mathbf{F}_k^{(1)} \right\|_2^2
\right),
\end{equation}
\begin{equation}
W_k = 1 + \alpha \cdot \mathbf{M} \cdot 
\min \left( 1, \frac{\text{iter}}{\text{total\_iter}} \right),
\end{equation}
where $L$ is the number of feature layers used for perceptual comparison, $\mathbf{F}_k^{(0)}$ and $\mathbf{F}_k^{(1)}$ denote the features extracted from the generated and ground-truth images at the $k_{th}$ layer, and $\mathcal{A}(\cdot)$ is an aggregation operator. The dynamic weight $W_k$ modulates the contribution of facial regions based on the binary mask $\mathbf{M}$, scaling factor $\alpha$, and a progressive training schedule.

This mechanism implements coarse-to-fine facial refinement: early training emphasizes global body structure, while later stages shift attention to fine-grained facial details. This leads to improved facial fidelity and perceptual realism, without compromising overall structural consistency.

\subsection{Layered Compression}
Following the hierarchical representation, we obtain pose information for each frame, including SMPL-X parameters, pose maps, and the StyleUNet network parameters that facilitate the mapping from poses to Gaussian parameters. For the three data types, we design corresponding encoding frameworks to achieve efficient compression and transmission.

\textbf{SMPL-X Parameter Compression.}  
SMPL-X parameters provide a compact and structured representation of human pose for each frame. Due to their high sensitivity to numerical precision, where even slight variations can result in noticeable errors in pose reconstruction, lossless compression is required to preserve accuracy. Given their limited value range and low redundancy, we apply Huffman coding to efficiently compress the SMPL-X parameters. This method preserves the structural fidelity of the pose while significantly reducing transmission costs.

\textbf{Pose Map Compression.}  
Pose maps represent the pose information of each frame in an image-like format. Given that each sequence is generated from the same individual, there is significant spatial redundancy between frames. To leverage this redundancy, we exploit temporal consistency for efficient encoding. We adopt DCVC-DC~\cite{li2023neural}, a deep learning-based video compression technique, which enhances context diversity both temporally and spatially, resulting in substantial coding gains. This method effectively preserves the fidelity of pose information while achieving high compression ratios, thereby reducing data size.

\begin{figure}[t]
    \centering
    \includegraphics[width=8.5cm]{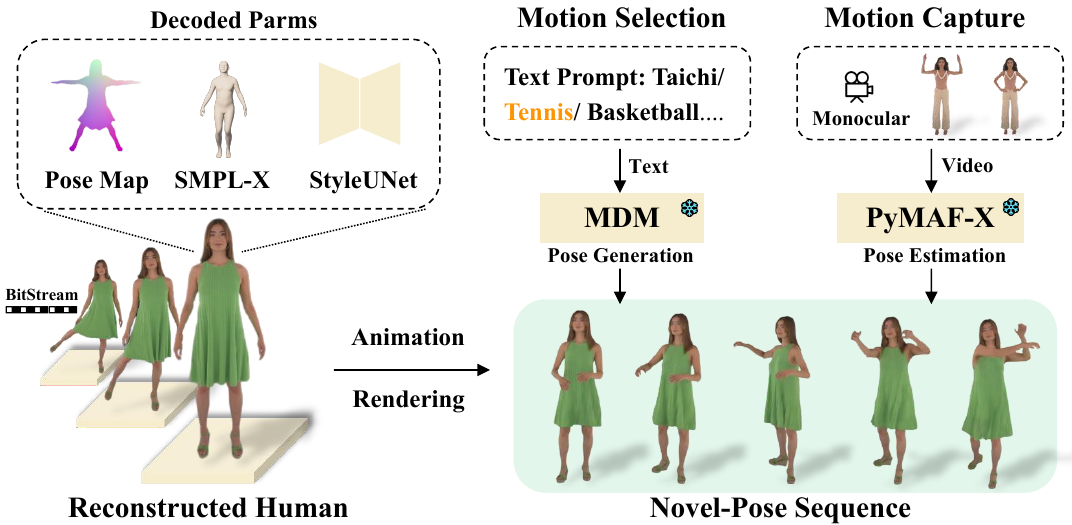}
    \caption{User motions are captured by edge devices or provided as motion pose text, enabling pose extraction or generation for novel pose synthesis in the reconstructed character.}
    \label{fig:animation}
\end{figure}

\textbf{StyleUNet Parameter Compression.}  
The StyleUNet network maps pose representations to Gaussian parameters but has significant transmission overhead due to its large size. To address this, we employ a quantization-based compression approach using greedy optimization~\cite{frantar2022optimal}. This method optimizes the quantization step size under a fixed bit-width constraint $\textit{Q}$, reducing redundancy while preserving the network’s expressive capability. A major advantage of this approach is its flexibility in bitrate control—by adjusting $\textit{Q}$, we can achieve compression at varying bitrates without retraining. This improves transmission efficiency and enhances adaptability in bandwidth-limited scenarios.

\subsection{GS-based Rendering and Interaction}\label{section:3.4}

We introduce a Gaussian-based rendering framework for efficient and accurate virtual avatar generation and interaction on edge devices. The framework consists of two main components: (1) decoding compressed data to retrieve the Gaussian maps (denoted as $G_f$ and $G_b$), which are then used for rendering high-quality 3D avatars, and (2) obtaining new poses from multimodal inputs (e.g., video or text) and interpolating to ensure that the generated pose maps are both visually consistent and realistic. The following sections elaborate on these two critical components.

\textbf{Gaussian Map Reconstruction and Rendering.}  
After decoding SMPL parameters and pose maps, StyleUNet generates per-frame Gaussian maps. Each pixel's Gaussian distribution encodes position, covariance, opacity, and color attributes, creating detailed 3D character representations across poses. To ensure complete coverage, we extract normalized 3D Gaussians from the predicted pose-related Gaussian map. While only front and back views are used during parameterization, orthogonal projections allow the resulting point cloud to cover additional areas, such as the sides and hands, providing sufficient information for realistic rendering.
For target pose rendering, we deform the normalized 3D Gaussians into pose space via LBS, which adjusts their positions and covariance attributes through rotation and translation operations derived from skinning weights. The deformed Gaussians are then rendered using splatting-based rasterization~\cite{kerbl20233Dgaussians} to generate the avatar image.

\textbf{Pose Driven and Interpolation. } 
We introduce a multimodal pose module that accepts both video and text inputs (Figure~\ref{fig:animation}). For video, PyMAF-X~\cite{zhang2023pymaf} extracts SMPL-X parameters, while text inputs are processed through a diffusion model~\cite{tevet2022human} to generate motion sequences subsequently converted to SMPL-X parameters.

To generate pose maps from these parameters, we employ PCA~\cite{mackiewicz1993principal} to project novel poses into the training pose space. The projection operates jointly on SMPL-X parameters and their corresponding training pose maps. During inference, new parameters are projected into this lower-dimensional space to generate the corresponding pose map, computed as:
\begin{equation}
    P(x) = S \cdot \max \left( \min \left( S^\top \left( P(x) - \mu \right), k\sigma \right), -k\sigma \right) + \mu ,
\end{equation}
where $P(x)$ denotes the input pose (from video/text), $S$ is the PCA matrix, $\mu$ is the mean of the training data, and $\sigma$ is the standard deviation of each component. The $\pm k\sigma$ clipping maintains plausible pose variations while preventing artifacts from extreme deviations.Finally, the pose maps are passed to the StyleUNet, which generates the corresponding Gaussian maps. Finally, we obtain the rendered human in the terminal. This outcome, when combined with the work on 3D scene generation~\cite{li2024scenedreamer360}, will contribute to supporting immersive communication and conferencing.

\begin{figure*}[htp]
    \centering
    \includegraphics[width=17.5cm]{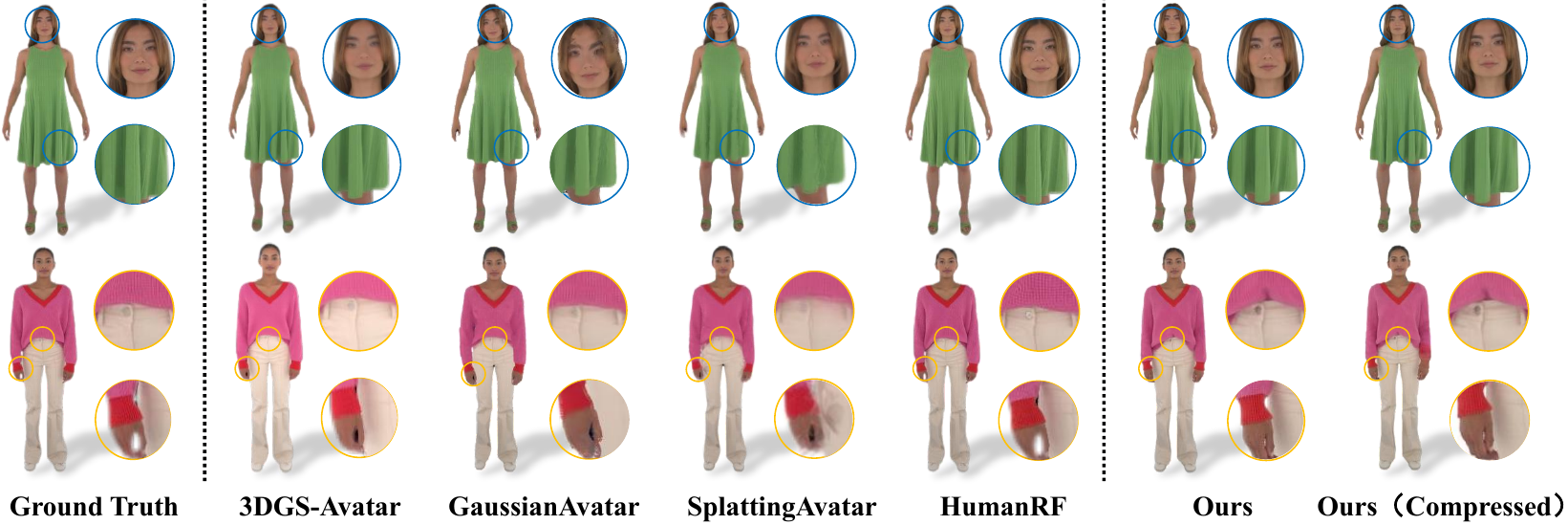}
    \caption{Comparison with reconstruction methods on ActorsHQ dataset~\cite{icsik2023humanrf}. Our method achieves a low bitrate while maintaining excellent reconstruction quality and texture details.}
    \label{fig:reconstruction}
\end{figure*}

\section{EXPERIMENT}
\subsection{Implementation Details}
Our HGC-Avatar is trained on a single NVIDIA L20 GPU using the Adam optimizer, with the core training module being the StyleUNet network. The learning rate was set to 0.0005, batch size to 1, and the loss function weights were configured as follows: $w_\text{L1}$ = 1.0, $w_\text{lpips}$ = 0.1, $w_\text{offset}$ = 0.005, and the facial module weight $\alpha$ = 0.2. The total number of iterations was $8.0 \times 10^5$. In the compression part, we achieve bitrate control through Q. Each experiment consumes an average of 10.3GB of GPU memory, and the training phase took approximately three days. With a rendering resolution of 1024x1024, we consume 1998MB of GPU memory, with a per-frame storage of 0.36MB and an average inference time of 0.11s.

\subsection{Datasets and Metrics}
\textbf{Datasets.}
To validate the high-quality reconstruction and low-bitrate compression performance of our framework, we conduct experiments on multiple datasets. These include two sequences from the ActorsHQ dataset~\cite{icsik2023humanrf}, three from AvatarRex~\cite{zheng2023avatarrex}, and three from THuman4.0~\cite{zheng2022structured}. The datasets contain about 2,000 frames each, captured from 16, 24, and 160 cameras, respectively. We use all video frames for training, and select 500-1000 frames for inference and comparison (500 for AvatarRex, 800 for THuman4.0, and 1000 for ActorsHQ). Data preprocessing, camera pose estimation, and other operations are performed before the experiments.

\noindent\textbf{Metrics.}
We evaluate both reconstruction quality and compression performance. Reconstruction is measured using PSNR, SSIM~\cite{wang2004image}, and LPIPS~\cite{zhang2018unreasonable}, comparing Gaussian-rendered results with ground truth. Compression is assessed through bitrate, estimating the average bitrate per frame for decoding. Additionally, we evaluate rendering time at the decoding end to assess inference speed during decoding in the application.
\setlength{\tabcolsep}{3pt}
\begin{table}
    \centering
    \small
    \caption{Reconstruction Performance Comparison on ActorsHQ~\cite{icsik2023humanrf} and AvatarRex~\cite{zheng2023avatarrex} datasets. We calculate metrics to measure reconstruction quality and model storage. We present the results before and after compression using our method.}
    \begin{tabular}{c|cccc}
    \toprule
        ActorsHQ & PSNR($\uparrow$) & SSIM($\uparrow$) & LPIPS($\downarrow$) & Storage ($\downarrow$) \\
    \midrule
        3DGS-Avatar~\cite{qian20243dgs} & 29.21 & 0.9535 & 0.0248 & 29.88MB  \\
       GaussianAvatar~\cite{hu2024gaussianavatar}& 23.20 & 0.9296 & 0.0420 & 2.99MB   \\
       SplattingAvatar~\cite{shao2024splattingavatar}& 23.89 & 0.9284 &  0.1176 & 11.49MB  \\
        HumanRF~\cite{icsik2023humanrf} & 30.13 & 0.9606 & 0.0432 & 2.05MB  \\
    \midrule
        Ours(Before) & 30.96 & 0.9708 & 0.0320 & 0.86MB \\
        Ours(After)& 29.96 & 0.9639 & 0.0343 & 0.32MB \\
    \toprule
        AvatarRex & PSNR($\uparrow$) & SSIM($\uparrow$) & LPIPS($\downarrow$) & Storage ($\downarrow$) \\
    \midrule
        3DGS-Avatar~\cite{qian20243dgs} & 28.14 & 0.9571 & 0.0552 & 7.17MB \\
       GaussianAvatar~\cite{hu2024gaussianavatar}&  23.04 & 0.9704 & 0.0306 & 2.44MB  \\
        SplattingAvatar~\cite{shao2024splattingavatar}& 25.95 & 0.9744 &  0.0613 & 17.23MB  \\
        GPS-Gaussian~\cite{zheng2024gps}& 28.86 & 0.9830 & 0.0090 & 51.61MB  \\
        AvatarRex~\cite{zheng2023avatarrex}& 23.70 & $-$ & 0.0440 & $-$ \\
    \midrule
        Ours(Before) & 30.42 & 0.9827 & 0.0257 & 1.73MB\\
        Ours(After)& 29.82 & 0.9815 & 0.0268 & 0.63MB \\
    \bottomrule
    \end{tabular}
    \label{tab:resonstruction1}
\end{table}

\subsection{Comparison with State-of-the-art Methods}
In this section, we compare reconstruction quality and compression performance. First, we compare our method with existing Gaussian-based human reconstruction approaches to demonstrate that our work can generate high-quality reconstructions with low bitrates. Then, we compare our method with several 3DGS compression techniques to validate its ability to achieve high-quality reconstruction under low-bitrate conditions. More reconstruction and driving results on the application side, as well as visual task applications, are presented in the supplementary materials.

\begin{figure*}[htp]
    \centering
    \includegraphics[width=17.5cm]{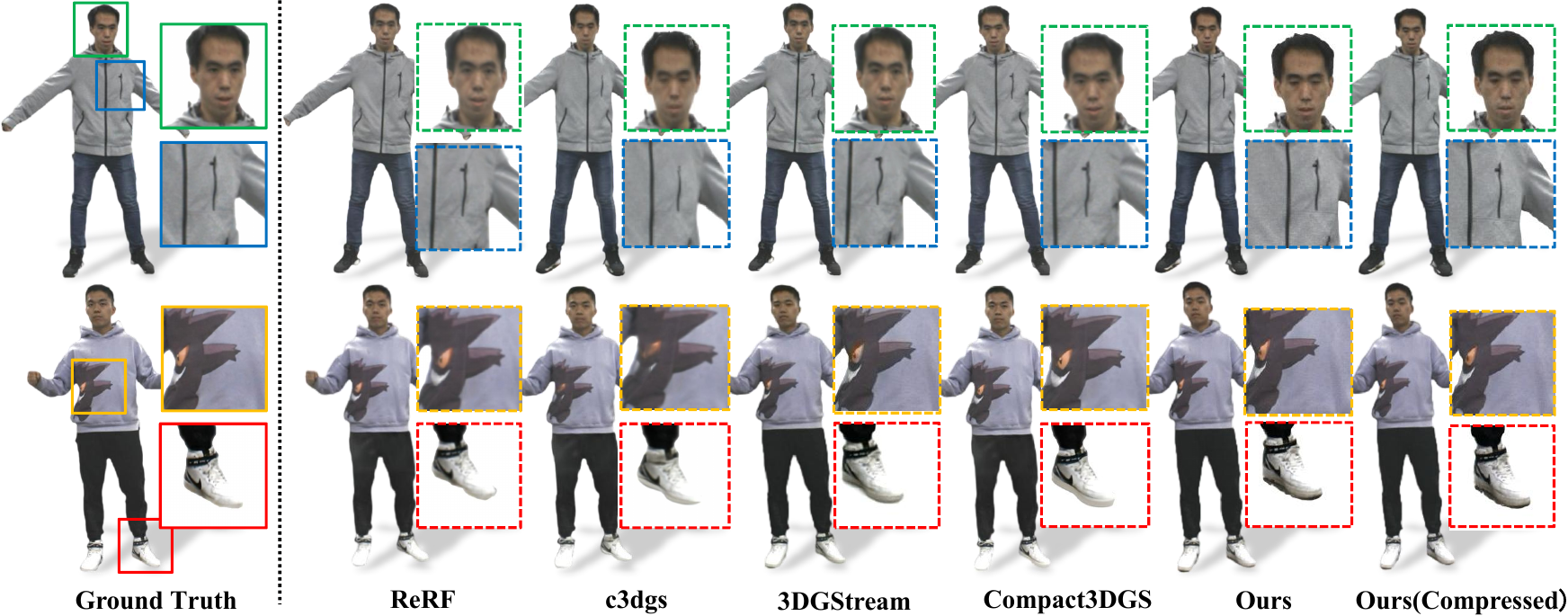}
    \caption{Comparison with compression methods on THuman4.0 dataset~\cite{zheng2022structured}. Our method maintains the best reconstruction quality even at low bitrates.}
    \label{fig:compression}
\end{figure*}

\noindent \textbf{Comparison with 3DGS Human Reconstruction.}
We compare our method with several 3DGS-based human reconstruction algorithms, including the monocular reconstruction methods 3DGS-Avatar~\cite{qian20243dgs}, GaussianAvatar~\cite{hu2024gaussianavatar}, and SplattingAvatar~\cite{shao2024splattingavatar}, as well as the multi-view input reconstruction methods GPS-Gaussian~\cite{zheng2024gps}, HumanRF~\cite{icsik2023humanrf}, and AvatarRex~\cite{zheng2023avatarrex}. Due to inconsistencies in dataset requirements and camera parameters for multi-view reconstruction methods, we conduct experimental comparisons only on specific datasets. Unfortunately, method AvatarRex~\cite{zheng2023avatarrex} does not provide open-source code, so we only report the reconstruction quality on the same dataset for reference.

The quantitative results are shown in Table~\ref{tab:resonstruction1}, and the subjective quality of reconstructed humans is displayed in Figure~\ref{fig:reconstruction}. We compare our method's pre-/post-compression performance with SOTA 3DGS approaches. While 3DGS-Avatar offers fast training/rendering and good quality, it suffers from a large model size. GaussianAvatar and SplattingAvatar are limited by monocular input, affecting reconstruction quality. HumanRF uses low-rank decomposition for high-fidelity 4D dynamic reconstructions with smaller model sizes, while GPS-Gaussian employs Gaussian parameter mapping for high-quality human reconstruction with computational efficiency. Our method outperforms or matches the best in PSNR, SSIM, and LPIPS scores. Thanks to an efficient hierarchical Gaussian representation, we achieve a low bitrate pre-compression and, with a multi-layer compression strategy, the lowest bitrate consumption, enabling efficient transmission and reconstruction.

\setlength{\tabcolsep}{3pt}
\begin{table}
    \centering
    \small
    \caption{Compression Performance Comparison on the THuman4.0
    dataset~\cite{zheng2022structured}. Under extremely low bitrates, our method can still maintain the highest reconstruction quality.}
    \begin{tabular}{c|ccccc}
    \toprule
        & PSNR($\uparrow$) & SSIM($\uparrow$) & LPIPS($\downarrow$) & Storage ($\downarrow$) & Time($\downarrow$)\\
    \midrule
        Compact3DGS~\cite{lee2024c3dgs} & 28.64 & 0.8124 & 0.2599 & 25.9MB  & 58min \\
       c3dgs~\cite{Niedermayr2024Compressed}&  28.89 & 0.8804 & 0.2496 & 28.7MB &  4min26s  \\
        ReRF~\cite{wang2023neural}& 32.25 & 0.9325 & 0.0498 & 5.93MB & 17min19s \\
        3DGStream~\cite{sun20243dgstream} & 31.35 & 0.9463 & 0.1769 & 6.3MB & 1h58min \\
    \midrule
        Ours(Before) & 33.74 & 0.9912 & 0.0290 & 1.08MB & 1min30s\\
        Ours(After)& 31.95 & 0.9864 & 0.0364 & 0.40MB & 1min30s \\
    \bottomrule
    \end{tabular}
    \label{tab:compression}
\end{table}

\noindent \textbf{Comparison with 3DGS Compression.}
Moreover, we compare our method with 3DGS compression approaches to verify its high reconstruction quality under low-bitrate transmission. Multiple experiments are conducted on the THuman4.0 dataset, comparing our approach with the current SOTA methods: Compact3DGS~\cite{lee2024c3dgs}, c3dgs~\cite{Niedermayr2024Compressed}, 3DGStream~\cite{sun20243dgstream} (which focuses more on streaming transmission), and the classic work ReRF~\cite{wang2023neural} for dynamic human compression. We calculate the reconstruction quality, bitrate, and inference time of different methods. Regarding storage calculation, we primarily compute the full grid for the first frame and inter-frame data thereafter. The experimental results are presented in Table~\ref{tab:compression} and Figure~\ref{fig:compression}. As shown, our method achieves the best SSIM and LPIPS scores after compression while maintaining the lowest bitrate and enabling fast inference rendering at the decoding end. Our approach demonstrates significant advantages in texture details and facial expressions.

\begin{figure}[htp]
    \centering
    \includegraphics[width=8.5cm]{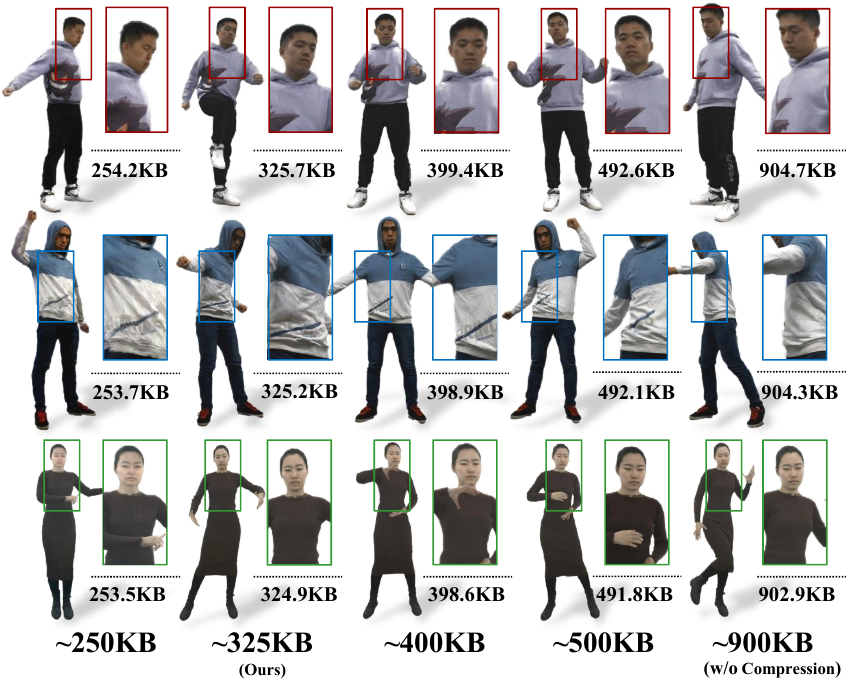}
    \caption{Ablation study on compression quantization step for AvatarReX~\cite{zheng2023avatarrex} and THuman4.0~\cite{zheng2022structured} shows reconstruction quality degrades with larger steps, with optimal balance achieved at ~325KB.}
    \label{fig:ablation-ratios}
\end{figure}

\begin{figure}[htp]
    \centering
    \includegraphics[width=8.5cm]{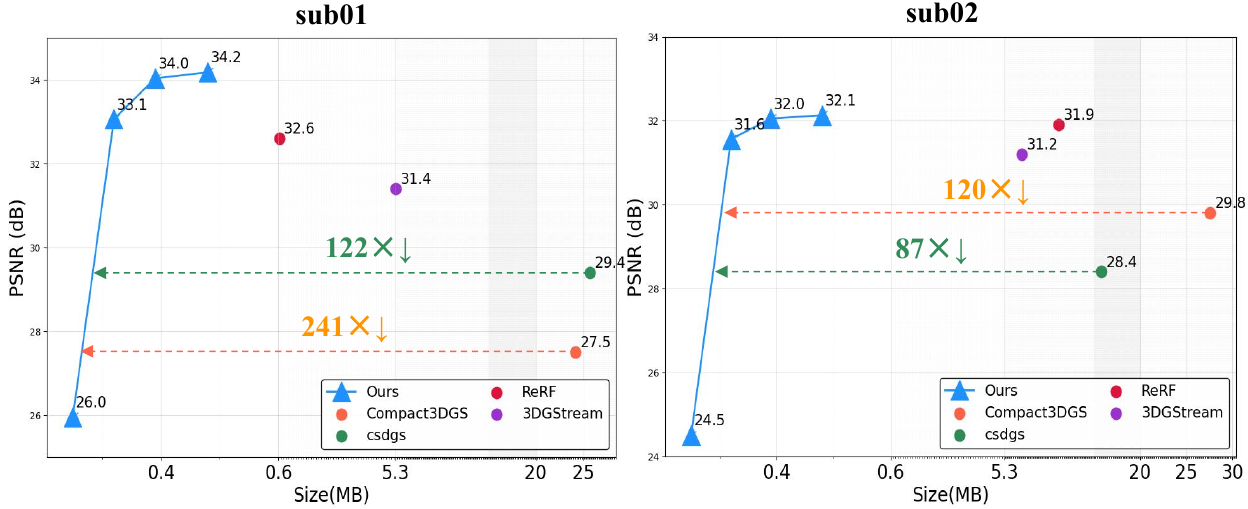}
    \caption{RD curves for quantitative comparisons. By adjusting quantization steps, we measure PSNR at various bitrates and compare with Gaussian compression approaches.}
    \label{fig:RDrate}
\end{figure}

\subsection{Evaluation}

\textbf{Quantization Step Analysis for Compression.}
The compressed parameters involve the bitrate points for video compression and the scaling factor for network compression. Under different bitrate controls, the reconstruction performance at the decoding end experiences varying degrees of degradation. The StyleUNet network accounts for the majority of the parameters, making the bitrate most affected by the network quantization step q. Keeping other parameters unchanged, we use networks with different quantization steps to drive 1000 frames in the sequence.

Figure~\ref{fig:ablation-ratios} shows the reconstruction results at different bitrates for the AvatarReX~\cite{zheng2023avatarrex} and THuman4.0~\cite{zheng2022structured} datasets. We visualize frames of human motion and observe that as the bitrate decreases, reconstruction quality deteriorates. The quantization step around 325KB represents the optimal balance point, beyond which quality drops significantly, impacting the visual experience. We also compare PSNR at various bitrates with other methods on the THuman4.0 dataset, with the Rate-Distortion (RD) curves shown in Figure~\ref{fig:RDrate}. The results demonstrate that our selected quantization step balances PSNR and bitrate effectively. Our method achieves about 100× compression compared to Gaussian methods while maintaining high reconstruction quality.

\noindent \textbf{Ablation on Facial Enhancement Module.}
In this section, we evaluate the facial enhancement module by removing the facial mask prior and running the same number of iterations on the THuman4.0 dataset. We compute reconstruction metrics and specifically assess facial fidelity by extracting the face region based on facial estimation. We calculate PSNR, SSIM~\cite{wang2004image}, and LPIPS~\cite{zhang2018unreasonable} for the face, along with the CLIP score~\cite{hessel2021clipscore} for semantic alignment and FID score~\cite{parmar2022aliased} for realism based on distributional similarity.

Table~\ref{tab:ablation} presents the ablation experiment results on multiple datasets. The first two rows in each group show results without face optimization, while the last two display final results. "All" indicates whole-body fidelity metrics, and "Face" focuses on facial quality. It can be observed that after facial enhancement, overall metrics such as PSNR and SSIM have significantly improved. Focusing on facial quality metrics, our method achieves more precise reconstruction of facial texture details, showing a more notable advantage in LPIPS, CLIP, and FID scores. More qualitative results are presented in the supplementary materials.

\begin{figure}[htp]
    \centering
    \includegraphics[width=8.5cm]{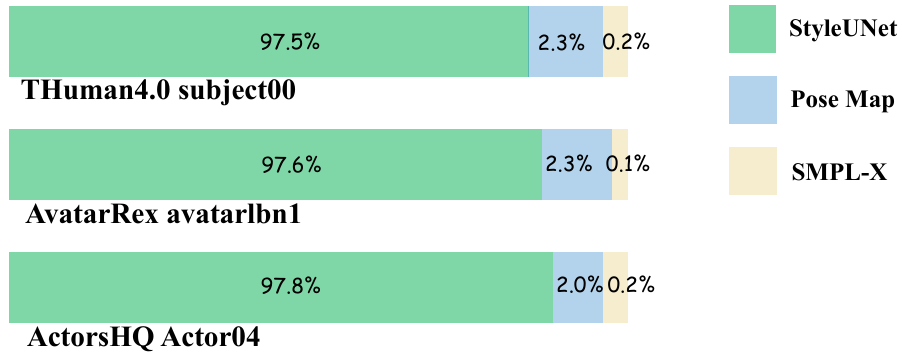}
    \caption{Bitstream Composition. The bitstream proportion of the encoded and decoded data for different datasets, representing the average weight of data from different layers.}
    \label{fig:code_part}
\end{figure}

\noindent \textbf{Bitstream Composition Structure.}
Figure~\ref{fig:code_part} shows our bitstream composition, consisting of SMPL-X parameters, Pose Map video, and StyleUNet parameters for motion and structure layers. The structure layer's network parameters dominate, as they map the pose map to Gaussian parameters, while the pose parameters and map require minimal bitstream.

\setlength{\tabcolsep}{3pt}
\begin{table}
    \centering
    \small
    \caption{Ablation Study on Face Enhancement on Thuman4.0 Dataset~\cite{zheng2022structured}. Our method significantly improves the reconstruction quality of facial details.}
    \begin{tabular}{c|c|ccccc}
    \toprule
        & Data Type & PSNR($\uparrow$) & SSIM($\uparrow$) & LPIPS($\downarrow$) & CLIP ($\uparrow$) & FID($\downarrow$)\\
    \midrule
        \multirow{4}{*}{sub00} & w/o face-All & 33.17 & 0.989 & 0.032 & 0.9470 & 24.08\\
        & w/o face-Face& 26.85 & 0.942 & 0.114 & 0.9118 & 34.90 \\
        & w/ face-All& 33.74 & 0.991 & 0.029 & 0.9490 & 23.61\\
        & w/ face-Face & 27.60 & 0.950 & 0.108 &  0.9170 & 32.71\\
    \midrule
        \multirow{4}{*}{sub02} & w/o face-All & 31.33 & 0.985 & 0.033 & 0.9385 & 26.31\\
        & w/o face-Face& 29.51 & 0.962 & 0.057 & 0.9230  & 30.66\\
        & w/ face-All& 32.13 & 0.987 & 0.030 & 0.9470 & 25.79\\
        & w/ face-Face & 29.93 & 0.967 & 0.055 & 0.9241 & 29.31\\
    \bottomrule
    \end{tabular}
    \label{tab:ablation}
\end{table}

\section{CONCLUSIONS}
In this paper, we propose a human-centric compression framework based on Gaussian Splatting for efficient representation. By disentangling structural geometry and temporal motion, our hierarchical approach enables efficient, controllable, and high-fidelity rendering of avatars. Leveraging a StyleUNet to map poses to Gaussian parameters, and encoding motion via compact SMPL representations, our method significantly reduces storage and transmission cost while enhancing rendering quality. The incorporation of facial attention further improves detail preservation in expressive regions. Experiments demonstrate that our approach achieves superior visual quality at lower bitrates compared to existing methods. Furthermore, our design supports motion editing at the receiver side, enabling flexible pose-driven rendering and precise avatar control. This work opens up new possibilities for high-speed rendering and deployment of dynamic 3D avatars on resource-constrained devices. Moreover, our approach provides valuable insights for future immersive communication and multi-viewpoint conferencing applications.

\begin{acks}
This work was supported by the National Key R\&D Program of China No. 2024YFB2809103, NSFC 62025101, BNSF No. L242014, PCL-CMCC Foundation for Science and Innovation Grant No. 2024ZY1C0040, CCF-Lenovo Open Funding 202301, Beijing Nova Program and New Cornerstone Science Foundation through the XPLORER PRIZE.
\end{acks}

\bibliographystyle{ACM-Reference-Format}
\balance
\bibliography{main}


\begin{thebibliography}{51}


\ifx \showCODEN    \undefined \def \showCODEN     #1{\unskip}     \fi
\ifx \showISBNx    \undefined \def \showISBNx     #1{\unskip}     \fi
\ifx \showISBNxiii \undefined \def \showISBNxiii  #1{\unskip}     \fi
\ifx \showISSN     \undefined \def \showISSN      #1{\unskip}     \fi
\ifx \showLCCN     \undefined \def \showLCCN      #1{\unskip}     \fi
\ifx \shownote     \undefined \def \shownote      #1{#1}          \fi
\ifx \showarticletitle \undefined \def \showarticletitle #1{#1}   \fi
\ifx \showURL      \undefined \def \showURL       {\relax}        \fi
\providecommand\bibfield[2]{#2}
\providecommand\bibinfo[2]{#2}
\providecommand\natexlab[1]{#1}
\providecommand\showeprint[2][]{arXiv:#2}

\bibitem[Chen et~al\mbox{.}(2024)]%
        {chen2024hac}
\bibfield{author}{\bibinfo{person}{Yihang Chen}, \bibinfo{person}{Qianyi Wu}, \bibinfo{person}{Weiyao Lin}, \bibinfo{person}{Mehrtash Harandi}, {and} \bibinfo{person}{Jianfei Cai}.} \bibinfo{year}{2024}\natexlab{}.
\newblock \showarticletitle{Hac: Hash-grid assisted context for 3d gaussian splatting compression}. In \bibinfo{booktitle}{\emph{European Conference on Computer Vision}}. Springer, \bibinfo{pages}{422--438}.
\newblock


\bibitem[Fan et~al\mbox{.}(2024)]%
        {fan2024lightgaussian}
\bibfield{author}{\bibinfo{person}{Zhiwen Fan}, \bibinfo{person}{Kevin Wang}, \bibinfo{person}{Kairun Wen}, \bibinfo{person}{Zehao Zhu}, \bibinfo{person}{Dejia Xu}, {and} \bibinfo{person}{Zhangyang Wang}.} \bibinfo{year}{2024}\natexlab{}.
\newblock \showarticletitle{LightGaussian: Unbounded 3D Gaussian Compression with 15x Reduction and 200+ {FPS}}. In \bibinfo{booktitle}{\emph{The Thirty-eighth Annual Conference on Neural Information Processing Systems}}.
\newblock
\urldef\tempurl%
\url{https://openreview.net/forum?id=6AeIDnrTN2}
\showURL{%
\tempurl}


\bibitem[Frantar and Alistarh(2022)]%
        {frantar2022optimal}
\bibfield{author}{\bibinfo{person}{Elias Frantar} {and} \bibinfo{person}{Dan Alistarh}.} \bibinfo{year}{2022}\natexlab{}.
\newblock \showarticletitle{Optimal brain compression: A framework for accurate post-training quantization and pruning}. In \bibinfo{booktitle}{\emph{Proc. Adv. Neural Inf. Process. Syst.}}, Vol.~\bibinfo{volume}{35}. \bibinfo{pages}{4475--4488}.
\newblock


\bibitem[Girish et~al\mbox{.}(2024)]%
        {girish2024eagles}
\bibfield{author}{\bibinfo{person}{Sharath Girish}, \bibinfo{person}{Kamal Gupta}, {and} \bibinfo{person}{Abhinav Shrivastava}.} \bibinfo{year}{2024}\natexlab{}.
\newblock \showarticletitle{Eagles: Efficient accelerated 3d gaussians with lightweight encodings}. In \bibinfo{booktitle}{\emph{European Conference on Computer Vision}}. Springer, \bibinfo{pages}{54--71}.
\newblock


\bibitem[Hessel et~al\mbox{.}(2021)]%
        {hessel2021clipscore}
\bibfield{author}{\bibinfo{person}{Jack Hessel}, \bibinfo{person}{Ari Holtzman}, \bibinfo{person}{Maxwell Forbes}, \bibinfo{person}{Ronan~Le Bras}, {and} \bibinfo{person}{Yejin Choi}.} \bibinfo{year}{2021}\natexlab{}.
\newblock \showarticletitle{Clipscore: A reference-free evaluation metric for image captioning}.
\newblock \bibinfo{journal}{\emph{arXiv preprint arXiv:2104.08718}} (\bibinfo{year}{2021}).
\newblock


\bibitem[Hu et~al\mbox{.}(2024)]%
        {hu2024gaussianavatar}
\bibfield{author}{\bibinfo{person}{Liangxiao Hu}, \bibinfo{person}{Hongwen Zhang}, \bibinfo{person}{Yuxiang Zhang}, \bibinfo{person}{Boyao Zhou}, \bibinfo{person}{Boning Liu}, \bibinfo{person}{Shengping Zhang}, {and} \bibinfo{person}{Liqiang Nie}.} \bibinfo{year}{2024}\natexlab{}.
\newblock \showarticletitle{Gaussianavatar: Towards realistic human avatar modeling from a single video via animatable 3d gaussians}. In \bibinfo{booktitle}{\emph{Proceedings of the IEEE/CVF conference on computer vision and pattern recognition}}. \bibinfo{pages}{634--644}.
\newblock


\bibitem[I{\c{s}}{\i}k et~al\mbox{.}(2023)]%
        {icsik2023humanrf}
\bibfield{author}{\bibinfo{person}{Mustafa I{\c{s}}{\i}k}, \bibinfo{person}{Martin R{\"u}nz}, \bibinfo{person}{Markos Georgopoulos}, \bibinfo{person}{Taras Khakhulin}, \bibinfo{person}{Jonathan Starck}, \bibinfo{person}{Lourdes Agapito}, {and} \bibinfo{person}{Matthias Nie{\ss}ner}.} \bibinfo{year}{2023}\natexlab{}.
\newblock \showarticletitle{Humanrf: High-fidelity neural radiance fields for humans in motion}.
\newblock \bibinfo{journal}{\emph{ACM Transactions on Graphics (TOG)}} \bibinfo{volume}{42}, \bibinfo{number}{4} (\bibinfo{year}{2023}), \bibinfo{pages}{1--12}.
\newblock


\bibitem[Jiang et~al\mbox{.}(2024a)]%
        {jiang2024robust}
\bibfield{author}{\bibinfo{person}{Yuheng Jiang}, \bibinfo{person}{Zhehao Shen}, \bibinfo{person}{Yu Hong}, \bibinfo{person}{Chengcheng Guo}, \bibinfo{person}{Yize Wu}, \bibinfo{person}{Yingliang Zhang}, \bibinfo{person}{Jingyi Yu}, {and} \bibinfo{person}{Lan Xu}.} \bibinfo{year}{2024}\natexlab{a}.
\newblock \showarticletitle{Robust dual gaussian splatting for immersive human-centric volumetric videos}.
\newblock \bibinfo{journal}{\emph{ACM Transactions on Graphics (TOG)}} \bibinfo{volume}{43}, \bibinfo{number}{6} (\bibinfo{year}{2024}), \bibinfo{pages}{1--15}.
\newblock


\bibitem[Jiang et~al\mbox{.}(2024b)]%
        {jiang2024hifi4g}
\bibfield{author}{\bibinfo{person}{Yuheng Jiang}, \bibinfo{person}{Zhehao Shen}, \bibinfo{person}{Penghao Wang}, \bibinfo{person}{Zhuo Su}, \bibinfo{person}{Yu Hong}, \bibinfo{person}{Yingliang Zhang}, \bibinfo{person}{Jingyi Yu}, {and} \bibinfo{person}{Lan Xu}.} \bibinfo{year}{2024}\natexlab{b}.
\newblock \showarticletitle{Hifi4g: High-fidelity human performance rendering via compact gaussian splatting}. In \bibinfo{booktitle}{\emph{Proceedings of the IEEE/CVF conference on computer vision and pattern recognition}}. \bibinfo{pages}{19734--19745}.
\newblock


\bibitem[Kerbl et~al\mbox{.}(2023)]%
        {kerbl20233Dgaussians}
\bibfield{author}{\bibinfo{person}{Bernhard Kerbl}, \bibinfo{person}{Georgios Kopanas}, \bibinfo{person}{Thomas Leimk{\"u}hler}, {and} \bibinfo{person}{George Drettakis}.} \bibinfo{year}{2023}\natexlab{}.
\newblock \showarticletitle{3D Gaussian Splatting for Real-Time Radiance Field Rendering}.
\newblock \bibinfo{journal}{\emph{ACM Transactions on Graphics}} \bibinfo{volume}{42}, \bibinfo{number}{4} (\bibinfo{date}{July} \bibinfo{year}{2023}).
\newblock
\urldef\tempurl%
\url{https://repo-sam.inria.fr/fungraph/3d-gaussian-splatting/}
\showURL{%
\tempurl}


\bibitem[Kumar et~al\mbox{.}(2024)]%
        {kumar2024styleunet}
\bibfield{author}{\bibinfo{person}{Karnamu~Naveen Kumar}, \bibinfo{person}{Aditya~Udaya Pattanaik}, {and} \bibinfo{person}{A~Robert Singh}.} \bibinfo{year}{2024}\natexlab{}.
\newblock \showarticletitle{StyleUNet: An Enhanced Style Transfer for Brain MRI Images using StyleGAN with U-Net}. In \bibinfo{booktitle}{\emph{2024 5th International Conference on Data Intelligence and Cognitive Informatics (ICDICI)}}. IEEE, \bibinfo{pages}{817--822}.
\newblock


\bibitem[Lee et~al\mbox{.}(2024)]%
        {lee2024c3dgs}
\bibfield{author}{\bibinfo{person}{Joo~Chan Lee}, \bibinfo{person}{Daniel Rho}, \bibinfo{person}{Xiangyu Sun}, \bibinfo{person}{Jong~Hwan Ko}, {and} \bibinfo{person}{Eunbyung Park}.} \bibinfo{year}{2024}\natexlab{}.
\newblock \showarticletitle{Compact 3D Gaussian Representation for Radiance Field}. In \bibinfo{booktitle}{\emph{Proceedings of the IEEE/CVF Conference on Computer Vision and Pattern Recognition (CVPR)}}. \bibinfo{pages}{21719--21728}.
\newblock


\bibitem[Li et~al\mbox{.}(2023)]%
        {li2023neural}
\bibfield{author}{\bibinfo{person}{Jiahao Li}, \bibinfo{person}{Bin Li}, {and} \bibinfo{person}{Yan Lu}.} \bibinfo{year}{2023}\natexlab{}.
\newblock \showarticletitle{Neural Video Compression with Diverse Contexts}. In \bibinfo{booktitle}{\emph{{IEEE/CVF} Conference on Computer Vision and Pattern Recognition, {CVPR} 2023, Vancouver, Canada, June 18-22, 2023}}.
\newblock


\bibitem[Li et~al\mbox{.}(2024a)]%
        {li2024scenedreamer360}
\bibfield{author}{\bibinfo{person}{Wenrui Li}, \bibinfo{person}{Fucheng Cai}, \bibinfo{person}{Yapeng Mi}, \bibinfo{person}{Zhe Yang}, \bibinfo{person}{Wangmeng Zuo}, \bibinfo{person}{Xingtao Wang}, {and} \bibinfo{person}{Xiaopeng Fan}.} \bibinfo{year}{2024}\natexlab{a}.
\newblock \showarticletitle{Scenedreamer360: Text-driven 3d-consistent scene generation with panoramic gaussian splatting}.
\newblock \bibinfo{journal}{\emph{arXiv preprint arXiv:2408.13711}} (\bibinfo{year}{2024}).
\newblock


\bibitem[Li et~al\mbox{.}(2024b)]%
        {li2024animatable}
\bibfield{author}{\bibinfo{person}{Zhe Li}, \bibinfo{person}{Zerong Zheng}, \bibinfo{person}{Lizhen Wang}, {and} \bibinfo{person}{Yebin Liu}.} \bibinfo{year}{2024}\natexlab{b}.
\newblock \showarticletitle{Animatable gaussians: Learning pose-dependent gaussian maps for high-fidelity human avatar modeling}. In \bibinfo{booktitle}{\emph{Proceedings of the IEEE/CVF conference on computer vision and pattern recognition}}. \bibinfo{pages}{19711--19722}.
\newblock


\bibitem[Liu et~al\mbox{.}(2024)]%
        {liu2024compgs}
\bibfield{author}{\bibinfo{person}{Xiangrui Liu}, \bibinfo{person}{Xinju Wu}, \bibinfo{person}{Pingping Zhang}, \bibinfo{person}{Shiqi Wang}, \bibinfo{person}{Zhu Li}, {and} \bibinfo{person}{Sam Kwong}.} \bibinfo{year}{2024}\natexlab{}.
\newblock \showarticletitle{Compgs: Efficient 3d scene representation via compressed gaussian splatting}. In \bibinfo{booktitle}{\emph{Proceedings of the 32nd ACM International Conference on Multimedia}}. \bibinfo{pages}{2936--2944}.
\newblock


\bibitem[Lu et~al\mbox{.}(2024)]%
        {lu2024scaffold}
\bibfield{author}{\bibinfo{person}{Tao Lu}, \bibinfo{person}{Mulin Yu}, \bibinfo{person}{Linning Xu}, \bibinfo{person}{Yuanbo Xiangli}, \bibinfo{person}{Limin Wang}, \bibinfo{person}{Dahua Lin}, {and} \bibinfo{person}{Bo Dai}.} \bibinfo{year}{2024}\natexlab{}.
\newblock \showarticletitle{Scaffold-gs: Structured 3d gaussians for view-adaptive rendering}. In \bibinfo{booktitle}{\emph{Proceedings of the IEEE/CVF Conference on Computer Vision and Pattern Recognition}}. \bibinfo{pages}{20654--20664}.
\newblock


\bibitem[Ma{\'c}kiewicz and Ratajczak(1993)]%
        {mackiewicz1993principal}
\bibfield{author}{\bibinfo{person}{Andrzej Ma{\'c}kiewicz} {and} \bibinfo{person}{Waldemar Ratajczak}.} \bibinfo{year}{1993}\natexlab{}.
\newblock \showarticletitle{Principal components analysis (PCA)}.
\newblock \bibinfo{journal}{\emph{Computers \& Geosciences}} \bibinfo{volume}{19}, \bibinfo{number}{3} (\bibinfo{year}{1993}), \bibinfo{pages}{303--342}.
\newblock


\bibitem[Mildenhall et~al\mbox{.}(2021)]%
        {Mildenhall2021NeRF}
\bibfield{author}{\bibinfo{person}{Ben Mildenhall}, \bibinfo{person}{Pratul~P. Srinivasan}, \bibinfo{person}{Matthew Tancik}, \bibinfo{person}{Jonathan~T. Barron}, \bibinfo{person}{Ravi Ramamoorthi}, {and} \bibinfo{person}{Ren Ng}.} \bibinfo{year}{2021}\natexlab{}.
\newblock \showarticletitle{NeRF: representing scenes as neural radiance fields for view synthesis}.
\newblock \bibinfo{journal}{\emph{Commun. ACM}} \bibinfo{volume}{65}, \bibinfo{number}{1} (\bibinfo{date}{Dec.} \bibinfo{year}{2021}), \bibinfo{pages}{99–106}.
\newblock
\showISSN{0001-0782}
\href{https://doi.org/10.1145/3503250}{doi:\nolinkurl{10.1145/3503250}}


\bibitem[Navaneet et~al\mbox{.}(2024)]%
        {Navaneet2024CompGS}
\bibfield{author}{\bibinfo{person}{K~L Navaneet}, \bibinfo{person}{Kossar Pourahmadi~Meibodi}, \bibinfo{person}{Soroush Abbasi~Koohpayegani}, {and} \bibinfo{person}{Hamed Pirsiavash}.} \bibinfo{year}{2024}\natexlab{}.
\newblock \showarticletitle{CompGS: Smaller and Faster Gaussian Splatting with Vector Quantization}. In \bibinfo{booktitle}{\emph{Computer Vision – ECCV 2024: 18th European Conference, Milan, Italy, September 29–October 4, 2024, Proceedings, Part XXXII}} (Milan, Italy). \bibinfo{publisher}{Springer-Verlag}, \bibinfo{address}{Berlin, Heidelberg}, \bibinfo{pages}{330–349}.
\newblock
\showISBNx{978-3-031-73410-6}
\href{https://doi.org/10.1007/978-3-031-73411-3_19}{doi:\nolinkurl{10.1007/978-3-031-73411-3_19}}


\bibitem[Newcombe et~al\mbox{.}(2015)]%
        {Newcombe2015Dynamic}
\bibfield{author}{\bibinfo{person}{Richard~A. Newcombe}, \bibinfo{person}{Dieter Fox}, {and} \bibinfo{person}{Steven~M. Seitz}.} \bibinfo{year}{2015}\natexlab{}.
\newblock \showarticletitle{DynamicFusion: Reconstruction and Tracking of Non-Rigid Scenes in Real-Time}. In \bibinfo{booktitle}{\emph{Proceedings of the IEEE Conference on Computer Vision and Pattern Recognition (CVPR)}}.
\newblock


\bibitem[Newcombe et~al\mbox{.}(2011)]%
        {Newcombe2011Kinect}
\bibfield{author}{\bibinfo{person}{Richard~A. Newcombe}, \bibinfo{person}{Shahram Izadi}, \bibinfo{person}{Otmar Hilliges}, \bibinfo{person}{David Molyneaux}, \bibinfo{person}{David Kim}, \bibinfo{person}{Andrew~J. Davison}, \bibinfo{person}{Pushmeet Kohi}, \bibinfo{person}{Jamie Shotton}, \bibinfo{person}{Steve Hodges}, {and} \bibinfo{person}{Andrew Fitzgibbon}.} \bibinfo{year}{2011}\natexlab{}.
\newblock \showarticletitle{KinectFusion: Real-time dense surface mapping and tracking}. In \bibinfo{booktitle}{\emph{2011 10th IEEE International Symposium on Mixed and Augmented Reality}}. \bibinfo{pages}{127--136}.
\newblock
\href{https://doi.org/10.1109/ISMAR.2011.6092378}{doi:\nolinkurl{10.1109/ISMAR.2011.6092378}}


\bibitem[Niedermayr et~al\mbox{.}(2024)]%
        {Niedermayr2024Compressed}
\bibfield{author}{\bibinfo{person}{Simon Niedermayr}, \bibinfo{person}{Josef Stumpfegger}, {and} \bibinfo{person}{R\"udiger Westermann}.} \bibinfo{year}{2024}\natexlab{}.
\newblock \showarticletitle{Compressed 3D Gaussian Splatting for Accelerated Novel View Synthesis}. In \bibinfo{booktitle}{\emph{Proceedings of the IEEE/CVF Conference on Computer Vision and Pattern Recognition (CVPR)}}. \bibinfo{pages}{10349--10358}.
\newblock


\bibitem[Parmar et~al\mbox{.}(2022)]%
        {parmar2022aliased}
\bibfield{author}{\bibinfo{person}{Gaurav Parmar}, \bibinfo{person}{Richard Zhang}, {and} \bibinfo{person}{Jun-Yan Zhu}.} \bibinfo{year}{2022}\natexlab{}.
\newblock \showarticletitle{On aliased resizing and surprising subtleties in gan evaluation}. In \bibinfo{booktitle}{\emph{Proceedings of the IEEE/CVF conference on computer vision and pattern recognition}}. \bibinfo{pages}{11410--11420}.
\newblock


\bibitem[Pavlakos et~al\mbox{.}(2019)]%
        {Pavlakos2019SMPL-X}
\bibfield{author}{\bibinfo{person}{Georgios Pavlakos}, \bibinfo{person}{Vasileios Choutas}, \bibinfo{person}{Nima Ghorbani}, \bibinfo{person}{Timo Bolkart}, \bibinfo{person}{Ahmed A.~A. Osman}, \bibinfo{person}{Dimitrios Tzionas}, {and} \bibinfo{person}{Michael~J. Black}.} \bibinfo{year}{2019}\natexlab{}.
\newblock \showarticletitle{Expressive Body Capture: 3D Hands, Face, and Body from a Single Image}. In \bibinfo{booktitle}{\emph{Proceedings IEEE Conf. on Computer Vision and Pattern Recognition (CVPR)}}.
\newblock


\bibitem[Peng et~al\mbox{.}(2021a)]%
        {peng2021animatable}
\bibfield{author}{\bibinfo{person}{Sida Peng}, \bibinfo{person}{Junting Dong}, \bibinfo{person}{Qianqian Wang}, \bibinfo{person}{Shangzhan Zhang}, \bibinfo{person}{Qing Shuai}, \bibinfo{person}{Xiaowei Zhou}, {and} \bibinfo{person}{Hujun Bao}.} \bibinfo{year}{2021}\natexlab{a}.
\newblock \showarticletitle{Animatable neural radiance fields for modeling dynamic human bodies}. In \bibinfo{booktitle}{\emph{Proceedings of the IEEE/CVF International Conference on Computer Vision}}. \bibinfo{pages}{14314--14323}.
\newblock


\bibitem[Peng et~al\mbox{.}(2021b)]%
        {peng2021neural}
\bibfield{author}{\bibinfo{person}{Sida Peng}, \bibinfo{person}{Yuanqing Zhang}, \bibinfo{person}{Yinghao Xu}, \bibinfo{person}{Qianqian Wang}, \bibinfo{person}{Qing Shuai}, \bibinfo{person}{Hujun Bao}, {and} \bibinfo{person}{Xiaowei Zhou}.} \bibinfo{year}{2021}\natexlab{b}.
\newblock \showarticletitle{Neural body: Implicit neural representations with structured latent codes for novel view synthesis of dynamic humans}. In \bibinfo{booktitle}{\emph{Proceedings of the IEEE/CVF conference on computer vision and pattern recognition}}. \bibinfo{pages}{9054--9063}.
\newblock


\bibitem[Pumarola et~al\mbox{.}(2021)]%
        {Pumarola2021DNerf}
\bibfield{author}{\bibinfo{person}{Albert Pumarola}, \bibinfo{person}{Enric Corona}, \bibinfo{person}{Gerard Pons-Moll}, {and} \bibinfo{person}{Francesc Moreno-Noguer}.} \bibinfo{year}{2021}\natexlab{}.
\newblock \showarticletitle{D-NeRF: Neural Radiance Fields for Dynamic Scenes}. In \bibinfo{booktitle}{\emph{Proceedings of the IEEE/CVF Conference on Computer Vision and Pattern Recognition (CVPR)}}. \bibinfo{pages}{10318--10327}.
\newblock


\bibitem[Qian et~al\mbox{.}(2024)]%
        {qian20243dgs}
\bibfield{author}{\bibinfo{person}{Zhiyin Qian}, \bibinfo{person}{Shaofei Wang}, \bibinfo{person}{Marko Mihajlovic}, \bibinfo{person}{Andreas Geiger}, {and} \bibinfo{person}{Siyu Tang}.} \bibinfo{year}{2024}\natexlab{}.
\newblock \showarticletitle{3dgs-avatar: Animatable avatars via deformable 3d gaussian splatting}. In \bibinfo{booktitle}{\emph{Proceedings of the IEEE/CVF conference on computer vision and pattern recognition}}. \bibinfo{pages}{5020--5030}.
\newblock


\bibitem[Shao et~al\mbox{.}(2024)]%
        {shao2024splattingavatar}
\bibfield{author}{\bibinfo{person}{Zhijing Shao}, \bibinfo{person}{Zhaolong Wang}, \bibinfo{person}{Zhuang Li}, \bibinfo{person}{Duotun Wang}, \bibinfo{person}{Xiangru Lin}, \bibinfo{person}{Yu Zhang}, \bibinfo{person}{Mingming Fan}, {and} \bibinfo{person}{Zeyu Wang}.} \bibinfo{year}{2024}\natexlab{}.
\newblock \showarticletitle{Splattingavatar: Realistic real-time human avatars with mesh-embedded gaussian splatting}. In \bibinfo{booktitle}{\emph{Proceedings of the IEEE/CVF Conference on Computer Vision and Pattern Recognition}}. \bibinfo{pages}{1606--1616}.
\newblock


\bibitem[Sun et~al\mbox{.}(2024a)]%
        {sun20243dgstream}
\bibfield{author}{\bibinfo{person}{Jiakai Sun}, \bibinfo{person}{Han Jiao}, \bibinfo{person}{Guangyuan Li}, \bibinfo{person}{Zhanjie Zhang}, \bibinfo{person}{Lei Zhao}, {and} \bibinfo{person}{Wei Xing}.} \bibinfo{year}{2024}\natexlab{a}.
\newblock \showarticletitle{3dgstream: On-the-fly training of 3d gaussians for efficient streaming of photo-realistic free-viewpoint videos}. In \bibinfo{booktitle}{\emph{Proceedings of the IEEE/CVF Conference on Computer Vision and Pattern Recognition}}. \bibinfo{pages}{20675--20685}.
\newblock


\bibitem[Sun et~al\mbox{.}(2024b)]%
        {sun2024f}
\bibfield{author}{\bibinfo{person}{Xiangyu Sun}, \bibinfo{person}{Joo~Chan Lee}, \bibinfo{person}{Daniel Rho}, \bibinfo{person}{Jong~Hwan Ko}, \bibinfo{person}{Usman Ali}, {and} \bibinfo{person}{Eunbyung Park}.} \bibinfo{year}{2024}\natexlab{b}.
\newblock \showarticletitle{F-3dgs: Factorized coordinates and representations for 3d gaussian splatting}. In \bibinfo{booktitle}{\emph{Proceedings of the 32nd ACM International Conference on Multimedia}}. \bibinfo{pages}{7957--7965}.
\newblock


\bibitem[Tevet et~al\mbox{.}(2022)]%
        {tevet2022human}
\bibfield{author}{\bibinfo{person}{Guy Tevet}, \bibinfo{person}{Sigal Raab}, \bibinfo{person}{Brian Gordon}, \bibinfo{person}{Yonatan Shafir}, \bibinfo{person}{Daniel Cohen-Or}, {and} \bibinfo{person}{Amit~H Bermano}.} \bibinfo{year}{2022}\natexlab{}.
\newblock \showarticletitle{Human motion diffusion model}.
\newblock \bibinfo{journal}{\emph{arXiv preprint arXiv:2209.14916}} (\bibinfo{year}{2022}).
\newblock


\bibitem[Wang et~al\mbox{.}(2023a)]%
        {wang2023neural}
\bibfield{author}{\bibinfo{person}{Liao Wang}, \bibinfo{person}{Qiang Hu}, \bibinfo{person}{Qihan He}, \bibinfo{person}{Ziyu Wang}, \bibinfo{person}{Jingyi Yu}, \bibinfo{person}{Tinne Tuytelaars}, \bibinfo{person}{Lan Xu}, {and} \bibinfo{person}{Minye Wu}.} \bibinfo{year}{2023}\natexlab{a}.
\newblock \showarticletitle{Neural residual radiance fields for streamably free-viewpoint videos}. In \bibinfo{booktitle}{\emph{Proceedings of the IEEE/CVF Conference on Computer Vision and Pattern Recognition}}. \bibinfo{pages}{76--87}.
\newblock


\bibitem[Wang et~al\mbox{.}(2024b)]%
        {wang2024videorf}
\bibfield{author}{\bibinfo{person}{Liao Wang}, \bibinfo{person}{Kaixin Yao}, \bibinfo{person}{Chengcheng Guo}, \bibinfo{person}{Zhirui Zhang}, \bibinfo{person}{Qiang Hu}, \bibinfo{person}{Jingyi Yu}, \bibinfo{person}{Lan Xu}, {and} \bibinfo{person}{Minye Wu}.} \bibinfo{year}{2024}\natexlab{b}.
\newblock \showarticletitle{Videorf: Rendering dynamic radiance fields as 2d feature video streams}. In \bibinfo{booktitle}{\emph{Proceedings of the IEEE/CVF Conference on Computer Vision and Pattern Recognition}}. \bibinfo{pages}{470--481}.
\newblock


\bibitem[Wang et~al\mbox{.}(2023b)]%
        {wang2023styleavatar}
\bibfield{author}{\bibinfo{person}{Lizhen Wang}, \bibinfo{person}{Xiaochen Zhao}, \bibinfo{person}{Jingxiang Sun}, \bibinfo{person}{Yuxiang Zhang}, \bibinfo{person}{Hongwen Zhang}, \bibinfo{person}{Tao Yu}, {and} \bibinfo{person}{Yebin Liu}.} \bibinfo{year}{2023}\natexlab{b}.
\newblock \showarticletitle{StyleAvatar: Real-time Photo-realistic Portrait Avatar from a Single Video}. In \bibinfo{booktitle}{\emph{ACM SIGGRAPH 2023 Conference Proceedings}}.
\newblock


\bibitem[Wang et~al\mbox{.}(2024c)]%
        {wang2024v}
\bibfield{author}{\bibinfo{person}{Penghao Wang}, \bibinfo{person}{Zhirui Zhang}, \bibinfo{person}{Liao Wang}, \bibinfo{person}{Kaixin Yao}, \bibinfo{person}{Siyuan Xie}, \bibinfo{person}{Jingyi Yu}, \bibinfo{person}{Minye Wu}, {and} \bibinfo{person}{Lan Xu}.} \bibinfo{year}{2024}\natexlab{c}.
\newblock \showarticletitle{V\^{} 3: Viewing Volumetric Videos on Mobiles via Streamable 2D Dynamic Gaussians}.
\newblock \bibinfo{journal}{\emph{ACM Transactions on Graphics (TOG)}} \bibinfo{volume}{43}, \bibinfo{number}{6} (\bibinfo{year}{2024}), \bibinfo{pages}{1--13}.
\newblock


\bibitem[Wang et~al\mbox{.}(2024a)]%
        {wang2024contextgs}
\bibfield{author}{\bibinfo{person}{Yufei Wang}, \bibinfo{person}{Zhihao Li}, \bibinfo{person}{Lanqing Guo}, \bibinfo{person}{Wenhan Yang}, \bibinfo{person}{Alex Kot}, {and} \bibinfo{person}{Bihan Wen}.} \bibinfo{year}{2024}\natexlab{a}.
\newblock \showarticletitle{Contextgs: Compact 3d gaussian splatting with anchor level context model}.
\newblock \bibinfo{journal}{\emph{Advances in neural information processing systems}}  \bibinfo{volume}{37} (\bibinfo{year}{2024}), \bibinfo{pages}{51532--51551}.
\newblock


\bibitem[Wang et~al\mbox{.}(2004)]%
        {wang2004image}
\bibfield{author}{\bibinfo{person}{Zhou Wang}, \bibinfo{person}{Alan~C Bovik}, \bibinfo{person}{Hamid~R Sheikh}, {and} \bibinfo{person}{Eero~P Simoncelli}.} \bibinfo{year}{2004}\natexlab{}.
\newblock \showarticletitle{Image quality assessment: from error visibility to structural similarity}.
\newblock  \bibinfo{volume}{13}, \bibinfo{number}{4} (\bibinfo{year}{2004}), \bibinfo{pages}{600--612}.
\newblock


\bibitem[Weng et~al\mbox{.}(2022)]%
        {weng2022humannerf}
\bibfield{author}{\bibinfo{person}{Chung-Yi Weng}, \bibinfo{person}{Brian Curless}, \bibinfo{person}{Pratul~P Srinivasan}, \bibinfo{person}{Jonathan~T Barron}, {and} \bibinfo{person}{Ira Kemelmacher-Shlizerman}.} \bibinfo{year}{2022}\natexlab{}.
\newblock \showarticletitle{Humannerf: Free-viewpoint rendering of moving people from monocular video}. In \bibinfo{booktitle}{\emph{Proceedings of the IEEE/CVF conference on computer vision and pattern Recognition}}. \bibinfo{pages}{16210--16220}.
\newblock


\bibitem[Xu et~al\mbox{.}(2024)]%
        {xu20244k4d}
\bibfield{author}{\bibinfo{person}{Zhen Xu}, \bibinfo{person}{Sida Peng}, \bibinfo{person}{Haotong Lin}, \bibinfo{person}{Guangzhao He}, \bibinfo{person}{Jiaming Sun}, \bibinfo{person}{Yujun Shen}, \bibinfo{person}{Hujun Bao}, {and} \bibinfo{person}{Xiaowei Zhou}.} \bibinfo{year}{2024}\natexlab{}.
\newblock \showarticletitle{4k4d: Real-time 4d view synthesis at 4k resolution}. In \bibinfo{booktitle}{\emph{Proceedings of the IEEE/CVF conference on computer vision and pattern recognition}}. \bibinfo{pages}{20029--20040}.
\newblock


\bibitem[Yan et~al\mbox{.}(2023)]%
        {yan2023model}
\bibfield{author}{\bibinfo{person}{Ruoke Yan}, \bibinfo{person}{Qian Yin}, \bibinfo{person}{Xinfeng Zhang}, {and} \bibinfo{person}{Siwei Ma}.} \bibinfo{year}{2023}\natexlab{}.
\newblock \showarticletitle{Model-Driven Compression for Digital Human Using Multi-Granularity Representations}. In \bibinfo{booktitle}{\emph{Proc. IEEE Int. Conf. Multimedia Expo}}. \bibinfo{pages}{690--695}.
\newblock
\href{https://doi.org/10.1109/ICME55011.2023.00124}{doi:\nolinkurl{10.1109/ICME55011.2023.00124}}


\bibitem[Yan et~al\mbox{.}(2024)]%
        {yan2024pose}
\bibfield{author}{\bibinfo{person}{Ruoke Yan}, \bibinfo{person}{Qian Yin}, \bibinfo{person}{Xinfeng Zhang}, \bibinfo{person}{Qi Zhang}, \bibinfo{person}{Gai Zhang}, {and} \bibinfo{person}{Siwei Ma}.} \bibinfo{year}{2024}\natexlab{}.
\newblock \showarticletitle{Pose-Driven Compression for Dynamic 3D Human via Human Prior Models}.
\newblock \bibinfo{journal}{\emph{IEEE Trans. Pattern Anal. Mach. Intell.}} \bibinfo{volume}{46}, \bibinfo{number}{8} (\bibinfo{year}{2024}), \bibinfo{pages}{5820--5834}.
\newblock
\href{https://doi.org/10.1109/TPAMI.2024.3368567}{doi:\nolinkurl{10.1109/TPAMI.2024.3368567}}


\bibitem[Yang et~al\mbox{.}(2024)]%
        {yang2024spectrally}
\bibfield{author}{\bibinfo{person}{Runyi Yang}, \bibinfo{person}{Zhenxin Zhu}, \bibinfo{person}{Zhou Jiang}, \bibinfo{person}{Baijun Ye}, \bibinfo{person}{Xiaoxue Chen}, \bibinfo{person}{Yifei Zhang}, \bibinfo{person}{Yuantao Chen}, \bibinfo{person}{Jian Zhao}, {and} \bibinfo{person}{Hao Zhao}.} \bibinfo{year}{2024}\natexlab{}.
\newblock \bibinfo{title}{Spectrally Pruned Gaussian Fields with Neural Compensation}.
\newblock
\showeprint[arxiv]{2405.00676}~[cs.CV]


\bibitem[Yariv et~al\mbox{.}(2021)]%
        {Yariv2021Volume}
\bibfield{author}{\bibinfo{person}{Lior Yariv}, \bibinfo{person}{Jiatao Gu}, \bibinfo{person}{Yoni Kasten}, {and} \bibinfo{person}{Yaron Lipman}.} \bibinfo{year}{2021}\natexlab{}.
\newblock \showarticletitle{Volume rendering of neural implicit surfaces} \emph{(\bibinfo{series}{NIPS '21})}. \bibinfo{publisher}{Curran Associates Inc.}, \bibinfo{address}{Red Hook, NY, USA}, Article \bibinfo{articleno}{367}, \bibinfo{numpages}{11}~pages.
\newblock
\showISBNx{9781713845393}


\bibitem[Zhang et~al\mbox{.}(2024)]%
        {zhang2024gaussianforest}
\bibfield{author}{\bibinfo{person}{Fengyi Zhang}, \bibinfo{person}{Yadan Luo}, \bibinfo{person}{Tianjun Zhang}, \bibinfo{person}{Lin Zhang}, {and} \bibinfo{person}{Zi Huang}.} \bibinfo{year}{2024}\natexlab{}.
\newblock \showarticletitle{GaussianForest: Hierarchical-Hybrid 3D Gaussian Splatting for Compressed Scene Modeling}.
\newblock \bibinfo{journal}{\emph{arXiv preprint arXiv:2406.08759}} (\bibinfo{year}{2024}).
\newblock


\bibitem[Zhang et~al\mbox{.}(2023)]%
        {zhang2023pymaf}
\bibfield{author}{\bibinfo{person}{Hongwen Zhang}, \bibinfo{person}{Yating Tian}, \bibinfo{person}{Yuxiang Zhang}, \bibinfo{person}{Mengcheng Li}, \bibinfo{person}{Liang An}, \bibinfo{person}{Zhenan Sun}, {and} \bibinfo{person}{Yebin Liu}.} \bibinfo{year}{2023}\natexlab{}.
\newblock \showarticletitle{Pymaf-x: Towards well-aligned full-body model regression from monocular images}.
\newblock \bibinfo{journal}{\emph{IEEE Transactions on Pattern Analysis and Machine Intelligence}} \bibinfo{volume}{45}, \bibinfo{number}{10} (\bibinfo{year}{2023}), \bibinfo{pages}{12287--12303}.
\newblock


\bibitem[Zhang et~al\mbox{.}(2018)]%
        {zhang2018unreasonable}
\bibfield{author}{\bibinfo{person}{Richard Zhang}, \bibinfo{person}{Phillip Isola}, \bibinfo{person}{Alexei~A Efros}, \bibinfo{person}{Eli Shechtman}, {and} \bibinfo{person}{Oliver Wang}.} \bibinfo{year}{2018}\natexlab{}.
\newblock \showarticletitle{The unreasonable effectiveness of deep features as a perceptual metric}. \bibinfo{pages}{586--595}.
\newblock


\bibitem[Zheng et~al\mbox{.}(2024)]%
        {zheng2024gps}
\bibfield{author}{\bibinfo{person}{Shunyuan Zheng}, \bibinfo{person}{Boyao Zhou}, \bibinfo{person}{Ruizhi Shao}, \bibinfo{person}{Boning Liu}, \bibinfo{person}{Shengping Zhang}, \bibinfo{person}{Liqiang Nie}, {and} \bibinfo{person}{Yebin Liu}.} \bibinfo{year}{2024}\natexlab{}.
\newblock \showarticletitle{Gps-gaussian: Generalizable pixel-wise 3d gaussian splatting for real-time human novel view synthesis}. In \bibinfo{booktitle}{\emph{Proceedings of the IEEE/CVF conference on computer vision and pattern recognition}}. \bibinfo{pages}{19680--19690}.
\newblock


\bibitem[Zheng et~al\mbox{.}(2022)]%
        {zheng2022structured}
\bibfield{author}{\bibinfo{person}{Zerong Zheng}, \bibinfo{person}{Han Huang}, \bibinfo{person}{Tao Yu}, \bibinfo{person}{Hongwen Zhang}, \bibinfo{person}{Yandong Guo}, {and} \bibinfo{person}{Yebin Liu}.} \bibinfo{year}{2022}\natexlab{}.
\newblock \showarticletitle{Structured Local Radiance Fields for Human Avatar Modeling}. In \bibinfo{booktitle}{\emph{Proceedings of the IEEE/CVF Conference on Computer Vision and Pattern Recognition (CVPR)}}.
\newblock


\bibitem[Zheng et~al\mbox{.}(2023)]%
        {zheng2023avatarrex}
\bibfield{author}{\bibinfo{person}{Zerong Zheng}, \bibinfo{person}{Xiaochen Zhao}, \bibinfo{person}{Hongwen Zhang}, \bibinfo{person}{Boning Liu}, {and} \bibinfo{person}{Yebin Liu}.} \bibinfo{year}{2023}\natexlab{}.
\newblock \showarticletitle{AvatarRex: Real-time Expressive Full-body Avatars}.
\newblock \bibinfo{journal}{\emph{ACM Transactions on Graphics (TOG)}} \bibinfo{volume}{42}, \bibinfo{number}{4} (\bibinfo{year}{2023}).
\newblock


\end{thebibliography}

\appendix
\clearpage

\end{document}